\documentclass[runningheads]{llncs}
\usepackage{graphicx}
\usepackage{amsmath,amssymb} 
\usepackage{color}
\usepackage{hyperref}
\usepackage[utf8]{inputenc}
\usepackage{tabulary}
\usepackage[width=154mm,left=28mm,paperwidth=210mm,height=219mm,top=30mm,paperheight=279mm]{geometry}
\begin{document}
\pagestyle{headings}
\setlength\extrarowheight{4pt}
\setlength{\tabcolsep}{4pt}
\renewcommand{\floatpagefraction}{.8}

\mainmatter

\title{End-to-End Interpretation of the French Street Name Signs Dataset}

\titlerunning{French Street Name Signs Dataset}

\authorrunning{Ray Smith et. al.}

\author{Raymond Smith, Chunhui Gu, Dar-Shyang Lee, Huiyi Hu, Ranjith Unnikrishnan, Julian Ibarz, Sacha Arnoud, Sophia Lin}


\institute{Google Inc, 1600 Ampthitheatre Pkwy, Mountain View, CA 94043, USA.\\
	\email{ \{rays,chunhui,dsl,clarahu,ranjith,julianibarz,sacha,sophi\}@google.com}
}

\maketitle

\begin{abstract}
We introduce the French Street Name Signs (FSNS) Dataset consisting of more than a million images of
 street name signs cropped from Google Street View images of France. Each image contains several views
 of the same street name sign. Every image has normalized, title case folded ground-truth text as it would
 appear on a map. 
 We believe that the FSNS dataset is large and complex enough to train a deep network of significant
 complexity to solve the street name extraction problem ``end-to-end" or to explore the design trade-offs
 between a single complex engineered network and multiple sub-networks designed and trained to solve
 sub-problems. We present such an ``end-to-end" network/graph for Tensor Flow and its results on the
 FSNS dataset.
\footnote{The final publication is available at link.springer.com:
\url{http://link.springer.com/chapter/10.1007/978-3-319-46604-0_30}}
\end{abstract}

\section{Introduction}

The detection and recognition of text from outdoor images is of increasing research interest to the fields
 of computer vision, machine learning and optical character recognition. The combination of perspective
 distortion, uncontrolled source text quality, and lack of significant structure to the text layout adds
 extra challenge to the still incompletely solved problem of accurately recognizing text from all the
 world's languages. Demonstrating the interest, several datasets related to the problem have become
 available: including ICDAR 2003 Robust Reading \cite{lucas2005icdar}, SVHN \cite{netzer2011reading},
 and, more recently, COCO-Text \cite{veit2016coco}, with details of these and others shown in
 Table~\ref{table:datasets}.

While these datasets each make a useful contribution to the field, the majority are very small compared
 to the size of a typical deep neural network. As the dataset size increases, it becomes increasingly
 difficult to maintain the accuracy of the ground-truth, as the task of annotating must be delegated
 to an increasingly large pool of workers less involved with the project.
 In the COCO-text \cite{veit2016coco} dataset for instance, the authors performed an audit themselves
 of the accuracy of the ground truth, and found that the annotators had found legible text regions
 with a recall of 84\%, and transcribed the text content with an accuracy of 87.5\%. Even at an edit
 distance of 1, the text content accuracy was still only 92.5\%, with missing punctuation being the
 largest remaining category of error.

Synthetic data has been shown \cite{Jaderberg14c} to be a good solution to this problem and can work well
provided the synthetic data generator includes the formatting/distortions that will be present in the
target problem. Some real-world data however, by its very nature, can be hard to predict, so real data
remains the first choice in many cases where available.

The difficulty remains therefore, in generating a sufficiently accurately annotated, large enough dataset
of real images, to satisfy the needs of modern data-hungry deep network-based systems, which can learn as
large a dataset as we can provide, without necessarily giving back the generalization that we would like.
To this end, and to make OCR more like image captioning, we present the French Street Name Signs (FSNS)
dataset, which we believe to be the first to offer multiple views of the same physical object, and thus
the chance for a learning system to compensate for degradation in any individual view.

\begin{table}
\begin{center}
\caption{Datasets of outdoor images containing text, including larger than single character ground truth. Information obtained mostly from the \href{http://www.iapr-tc11.org/mediawiki/index.php/Datasets}{iapr-tc11.org} website}
\label{table:datasets}
{\scriptsize
\begin{tabulary}{\linewidth}{ | C | C | C |}
\hline
Name & Content & Size \\ \hline \hline 
ICDAR2003 \cite{lucas2005icdar} & Images with word and character bounding boxes & Train: 258 Images, 1,157 words Test: 251 Images, 1,111 words \\ \hline
SVHN \cite{netzer2011reading} & Images of numbers and single digits from Google Street View with boxes & Train: 73,257 digits Test: 26,032 Additional: 531,131 \\ \hline
COCO-text \cite{veit2016coco} & Images from the MS COCO dataset that contain text & 63,686 images with 173,589 text regions \\ \hline
KAIST \cite{jung2011touch} Scene Text & Images with word and character boxes of Korean and English & 3,000 images\\ \hline 
NEOCR \cite{nagy2011neocr} & Images with text field boxes and perspective quadrangles. & 659 images with 5,238 text fields\\ \hline 
SVT \cite{wang2011end} & Images from Google Street View, with names of businesses in them & Train: 100 images, 211 words Test: 250 images, 514 words \\ \hline
Synthetic Word \cite{Jaderberg14c} & Synthetic images of real-world-like words & 9 million images, 90k distinct words \\ \hline
FSNS & Images of French street name signs & \textgreater 1,000,000 images \\ \hline
\end{tabulary}
}
\end{center}
\end{table}
\section{Basics of the FSNS Dataset}
\begin{figure}
\centering
\includegraphics[height=2.5cm]{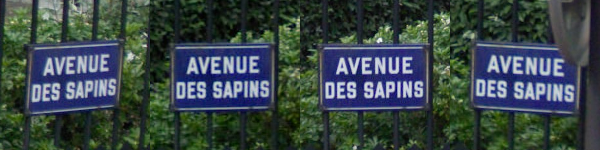}
\includegraphics[height=2.5cm]{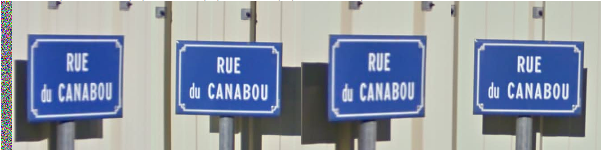}
\includegraphics[height=2.5cm]{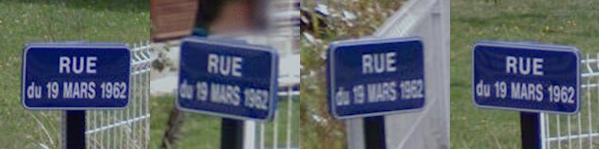}
\caption{Some examples of FSNS images}
\label{fig:samples}
\end{figure}

As its name suggests, the FSNS dataset is a set of signs, from the streets of France, that bear street
 names. Some example images are shown in Figure~\ref{fig:samples}. Each image carries four tiles of
 $150 \times 150$ pixels laid out horizontally, each of which contains a pre-detected street name sign,
 or random noise in the case that less than four independent views are available of the same physical
 sign. The text detection problem is thus largely eliminated, although the signs are still of variable
 size and orientation within each tile image. Also each sign carries multiple text lines, with a maximum
 of 3 lines of significant text, with the possibility of other additional lines of irrelevant text. Each
 of the tiles within an image is intended to be a different view {\it of the same physical sign,} taken
 from a different position and/or at a different time. Different physical signs of the same street name,
 from elsewhere on the same street, are included as separate images. There are over 1 million different
 physical signs.

\begin{figure}
\centering
\includegraphics[height=2.5cm]{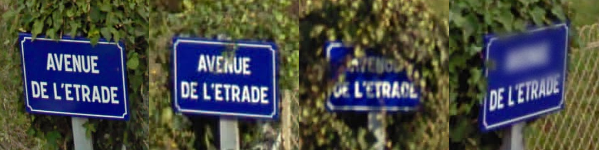}
\includegraphics[height=2.5cm]{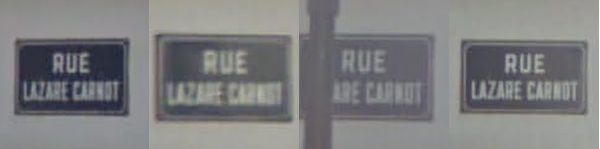}
\includegraphics[height=2.5cm]{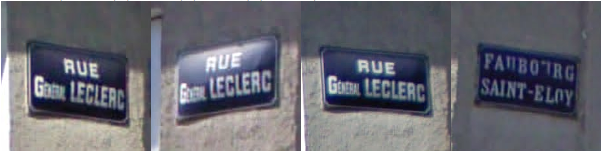}
\caption{Examples of blurring, obstruction, and incorrect spatial clustering}
\label{fig:blurring}
\end{figure}

The different views are of different quality, possibly taken from an acute angle, or blurred by motion,
 distance from the camera, or by unintentional privacy filtering. Occasionally some of the tiles may
 be views of a different sign altogether, which can happen when two signs are attached to the same post.
 Some examples of these problems are shown in Figure~\ref{fig:blurring}. The multiple views can reduce
 some of the usual problems of outdoor images, such as occlusion by foreground objects, image truncation
 caused by the target object being at the edge of the frame, and varied lighting. Other problems cannot
 be solved by multiple views, such as bent, corroded or faded signs.

The task of the system then is to obtain the best possible canonical text result by combining information from the multiple views, either by processing each tile independently and combining the results, or by combining information deep within the recognition system (most likely deep network).

\section{How the FSNS Dataset Was Created}
\label{sec:created}

The following process was used to create the FSNS dataset:
\begin{enumerate}
  \item A street-name-sign detector was applied to all Google Street View images from France. The detector returns an image rectangle around each street name sign, together with its geographic location (latitude and longitude).
  \item Multiple images of the same geographic location were gathered together (spatially clustered).
  \item Text from the signs was transcribed using a combination of reCAPTCHA \cite{recaptcha}, OCR and human operators.
  \item Transcribed text was presented to human operators to verify the accuracy of the transcription. Incorrect samples were re-routed for human transcription (back to step 3) or discarded if already the result of a human transcription.
  \item Images were bucketized geographically (by latitude/longitude) so that the train, validation, test,
 and private test sets come from disjoint geographic locations, with 100 m wide strips of ``wall" in
 between that are not used, to ensure that the same physical sign can't be viewed from different sets.
  \item Since roads are long entities that may pass between the disjoint geographic sections, there may be multiple signs of the same street name at multiple locations in different subsets. Therefore as each subset is generated, any images with truth strings that match a truth string in any previously generated subset are discarded. Each subset thereby has a disjoint set of truth strings.
  \item All images for which the truth string included a character outside of the chosen encoding set, or for which the encoded label length exceeded the maximum of 37, were discarded. The character set to be handled is thus carefully controlled.
\end{enumerate}

Note that the transcription was systematically Title Case folded from the original transcription, in order
 to make it represent the way that the street name would appear on a map. This process includes removal of
 text that is not relevant, including data such as the district or building numbers.

\section{Normalized Truth Text} \label{normalizedtruth}
The FSNS dataset is made more interesting by the fact that the truth text is a normalized representation
 of the name of the street, as it should be written on the map, instead of a simple direct transcription
 of the text on the sign. The main normalization is Title Case transformation of the text, which is often
 written on the sign in all upper case. Title Case is specified as follows:
\begin{quote}
The words: au, aux, de, des, du, et, la, le, les, sous, sur  always appear in lower-case.
The prefixes: d', l' always appear in lower-case.
All other words, including suffixes after d' and l', always appear with the initial letter capitalized and the rest in lower-case.
\end{quote}
The other main normalization is that some text on the sign, which is not part of the name of the street, is discarded. Although this seems a rather vague instruction, for a human, even without knowledge of French, it becomes easy after reading a few signs, as the actual street names fit into a reasonably obvious pattern, and the extraneous text is usually in a smaller size.

\begin{figure}
\centering
\includegraphics[height=2.5cm]{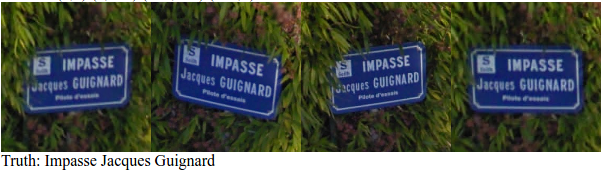}
\includegraphics[height=2.5cm]{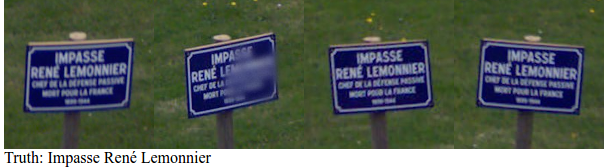}
\includegraphics[height=2.5cm]{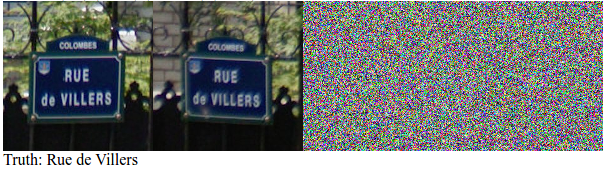}
\includegraphics[height=2.5cm]{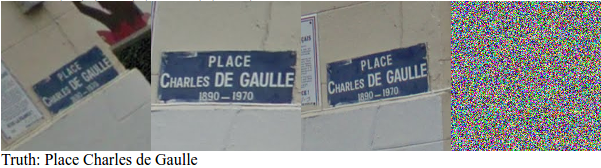}
\caption{Examples of images with their normalized truth text}
\label{fig:normalized}
\end{figure}

Some examples of some of these normalizations of the text between the sign and the truth text are shown in
 Figure~\ref{fig:normalized}.
 The task of transcribing the signs is thus not a basic OCR problem, but perhaps somewhat more like image
 captioning \cite{vinyals2015show}, by requiring an interpretation of what the sign {\it means,} not just its
 literal content. A researcher working with the FSNS dataset is hereby provided with a variety of design
 options between adding text post-processing to the output of an OCR engine and training a single
 network to learn the entire problem ``end-to-end".

\section{Details of the FSNS Dataset}
The location of the FSNS dataset is documented in the \texttt{README.md} file.\footnote{\url{https://github.com/tensorflow/models/tree/master/street/README.md}}
 There are 3 disjoint subsets, Train, Validation and Test\footnote{An additional private test set will
 be kept back for the purposes of organizing competitions.}. Each contains images of fixed size,
 $600\times150$ pixels, containing 4 tiles of $150\times150$ laid out horizontally, and padded with random noise where less than 4 views are available.

The size and location of each subset are shown in Table~\ref{table:sizes}, and some basic analysis of the word content of each subset is shown in Table~\ref{table:words}. The analysis in Table~\ref{table:words} excludes frequent words with frequency in the Train set \textgreater 100, and the words listed in Section ~\ref{normalizedtruth} as lower-case. As might be expected, given the process by which the subsets have been made disjoint, the fraction of words in each subset that are out of vocabulary with respect to the Train subset is reasonably high at around 30\%. Such a rate of out-of-vocabulary words will also make it difficult for a system to learn the full vocabulary from the Train set.

\begin{table}
\begin{center}
\caption{Location and size of each subset of the FSNS dataset}
\label{table:sizes}
{\scriptsize
\begin{tabulary}{\linewidth}{| C | C | C | C |}
\hline
Subset 	     & Location				   & Number of Images & Number of Words \\ \hline \hline
Train 	     & train/train@512 	   & 1044868 & 3189576 \\ \hline
Validation   & validation/validation@64 & 16150 & 50218 \\ \hline
Test 	     & test/test@64 		   & 20404 & 62650 \\ \hline
Private Test & n/a 				   & 21054 & 65366 \\ \hline
\end{tabulary}
}
\end{center}
\end{table}

\begin{table}
\begin{center}
\caption{Word counts excluding `stop' words, (being the prefixes with a frequency \textgreater 100, and the lower-cased words) in each subset and number out of vocabulary (OOV) with respect to (wrt) words in the Train subset. }
\label{table:words}
{\scriptsize
\begin{tabulary}{\linewidth}{ | c | C | C | C | C | C |}
\hline
Subset       & Non-stop Words & Unique Words & Unique Words OOV wrt Train & Total OOV words    & Percent OOV words \\ \hline \hline
Train        & 1336341     & 93482        & 0             & 0         & 0 \\ \hline
Validation     & 22250      & 7425        & 3482          & 7272     & 32.7 \\ \hline
Test           & 28587        & 8675   & 4081        & 8526     & 29.8 \\ \hline
Private Test   & 28752      & 8870        & 4265          & 9375     & 32.6 \\ \hline
\end{tabulary}
}
\end{center}
\end{table}

Each subset is stored as multiple TFRecords files of \texttt{tf.train.Example} protocol buffers, which makes them ready-made
 for input to TensorFlow \cite{tensorflow}\cite{abadi2016tensorflow}.
 The Example protocol buffer is very flexible, so the full details of the content of each example are laid
 out in Table~\ref{table:tfspec}.

Note that the ultimate goal of a machine learning system is to produce the UTF-8 string in ``image/text."
 That may be achieved simply by learning the byte sequences in the text field. Alternatively, there is also a
 pre-encoded mapping to integer class-ids provided in ``image/class" and ``image/unpadded\_class".
 The mapping between these class-ids and the UTF-8 text is provided in a separate file entitled
 \texttt{charset\_size=134.txt.} Each line in that file lists a class-id, a tab character, and the UTF-8 string
 that is represented by the class-id. Class-id 0 represents a space, and the last class-id, 133,
 represents the ``null" character, as used by the Connectionist Temporal Classification (CTC) alignment
 algorithm \cite{graves2006connectionist} typically used with an LSTM network. Note that some
 class-ids map to multiple UTF-8 strings, as some normalization has been applied, such as folding all the different shapes of double quote to the same class.

The ground truth text in the FSNS dataset uses a subset of these characters. In addition to all digits,
upper and lower-case A-Z, there are the following accented characters:
à
À
â
Â
ä
ç
Ç
é
É
è
È
ê
Ê
ë
Ë
î
Î
ï
ô
Ô
œ
ù
Ù
û
Û
ü
ÿ
and these punctuation symbols:
\textless
=
\_
-
,
;
!
?
/
.
'
"
(
)
]
\
\&
+
 a total of 109, including space.

For systems that process the multiple views separately, it is possible to avoid processing the noise padding. The number of real, non-noise views of a sign is given by the value of
the field ``image/orig\_width" divided by 150.

No sample in any of the subsets has a text field that encodes to more than 37 class-ids.
 37 is not a completely arbitrary choice. When padded with nulls in between each label for CTC,
 $(2\times37+1=75)$ the classid sequences are no longer than half the width $(150/2=75)$ of a single
 input view, which allows for some shrinkage of the data width in the network.

\begin{table}
\begin{center}
\caption{The content of each Example proto in the TFRecords files}
\label{table:tfspec}
{\scriptsize
\begin{tabulary}{\linewidth}{| c | c | c | C |}
\hline
Key name		& Type 		& Length   & Content \\ \hline \hline
image/format  	& bytes(string) & 1 	   & ``PNG" \\ \hline 
image/encoded 	& bytes(string) & 1	   & Image encoded as PNG. \\ \hline 
image/class   	& int64		& 37	   & Truth class-ids padded with nulls. \\ \hline
image/unpadded\_class & int64		& Variable & Truth class-ids unpadded. \\ \hline
image/width		& int64		& 1	   & Width of the image in pixels. \\ \hline
image/orig\_width	& int64		& 1  	   & Pre-padding width in pixels. \\ \hline 
image/height		& int64		& 1	   & Height of the image in pixels. \\ \hline
image/text		& bytes(string)	& 1	   & Truth string in UTF-8.\\ \hline
\end{tabulary}
} 
\end{center} 
\end{table}

\section{The Challenge}\label{challenge}

The FSNS dataset provides a rich and interesting challenge in machine learning, due to the variety of
 tasks that are required. Here is a summary of the different processes that a model needs to learn to
 discover the right solution:
\begin{itemize}
\setlength\itemsep{0.5em}
\item[\labelitemii] Locating the lines of text within the sign within each image.
\item[\labelitemii] Recognizing the text content within each line.
\item[\labelitemii] Discarding irrelevant text.
\item[\labelitemii] Title Case normalization.
\item[\labelitemii] Combining data from multiple signs, ignoring data from blurred or inconsistent signs.
\end{itemize}

None of the above is an explicit goal of the challenge. The current trend in machine learning is
 to build and train a single large/deep network to solve all of a problem without additional
 algorithmic pieces on one end or another, or to glue trained components
 together \cite{vinyals2015show}\cite{graves2014towards}. We believe that the FSNS data set is large enough to train
 a single deep network to learn all of the above tasks,
 and we provide an example in Section \ref{baseline}.
 We therefore propose that a competition based on the FSNS dataset should measure:
\begin{itemize}
\setlength\itemsep{0.5em}
\item[\labelitemii] Word recall: Fraction of space-delimited words in the truth that are present in the OCR output.
\item[\labelitemii] Word precision: Fraction of space-delimited words in the OCR output that are present in the truth.
\item[\labelitemii] Sequence error: the fraction of truth text strings that are not produced exactly by
 the network, after folding multiple spaces to single space.
\end{itemize}
Word recall and precision are almost universally used, and need no introduction. We add sequence error
 here because the strings are short enough that we can expect a significant number of them to be
 completely correct. Using only these metrics allows for end-to-end systems to compete directly against
systems built from smaller components that are designed for specific sub-problems.

\section{An End-to-End Solution} \label{baseline}

\begin{figure}
\centering
\vspace{-8cm}
\includegraphics[width=\textwidth]{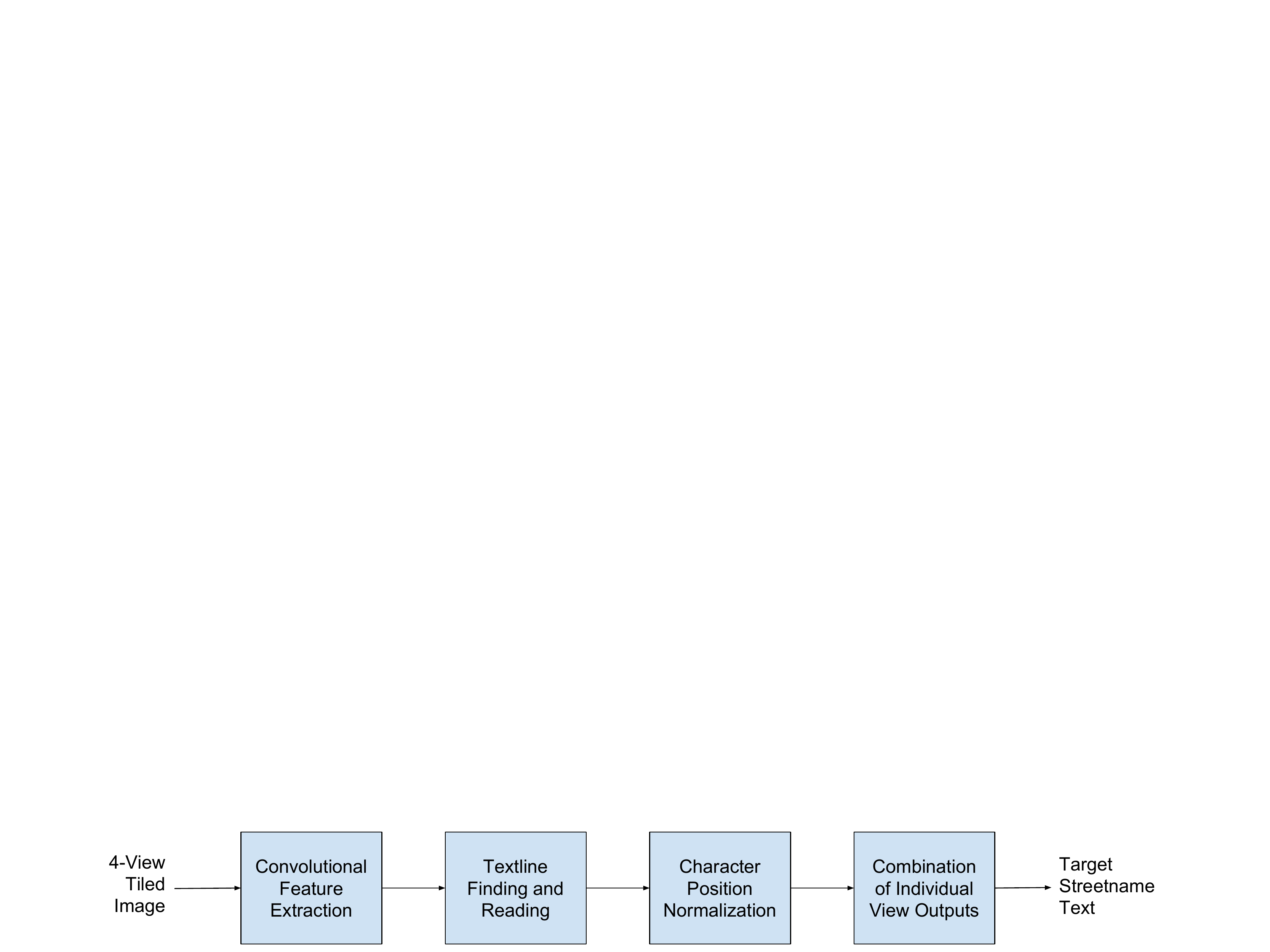}
\caption{High-level structure of the network graph}
\label{fig:net}
\end{figure}

We now describe a Tensor Flow graph that has been designed specifically to address the Challenge,
 end-to-end, using just the graph, with no algorithmic components. This means that the text line
finding and handling of multiple views, including where there are less than four, is entirely learned
and dealt with inside the network. Instead of using the orig\_width field in the dataset, the images
are input as fixed size and the random padding informs the network of the lack of useful content.
 The network is based on the design that has been shown to work well for many languages in
 Tesseract \cite{tutorial},
 with some extensions to handle the multi-line, multi-tile FSNS dataset.
 The design is named Street-name Tensor-flow Recurrent End-to-End Transcriber (STREET).
 To perform the tasks listed above, the graph design has a high-level structure with purpose,
 as shown in Figure~\ref{fig:net}.

Conventional convolutional layers process the images to extract features. Since each view may contain
 up to three lines of text, the next step is intended to allow the network to find upto three text lines
and recognize the text in each separately.
 The text may appear in different positions within each image, so some character
 position normalization is also required. Only then can the individual outputs be combined to produce
 a single target string. These components of the end-to-end system are described in detail below. 
Tensor Flow code for the STREET model described in this paper is available at the Tensor Flow Github
repository.\footnote{\url{https://github.com/tensorflow/models/tree/master/street}} 

\subsection{Convolutional Feature Extraction}

\begin{figure}
\centering
\vspace{-4.5cm}
\includegraphics[width=\textwidth]{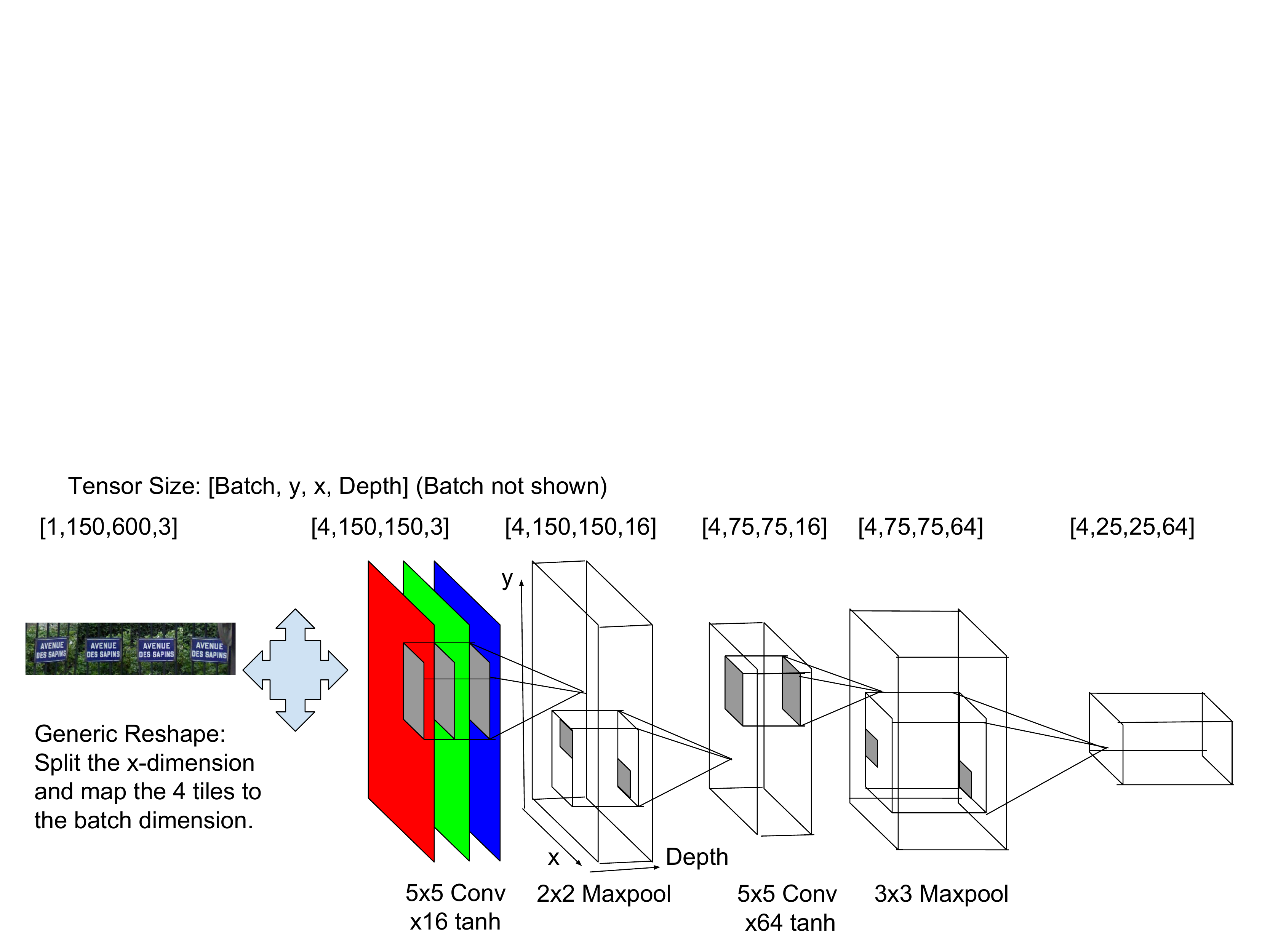}
\caption{Convolutional Feature Extraction and size reduction}
\label{fig:conv}
\end{figure}

The input image, being $600\times150$, is de-tiled to make the input a batch of 4 images of size $150\times150$.
 This is achieved by a generic reshape, which is a combination of TensorFlow reshape and transpose
 operations that split one dimension of the input tensor and map the split parts to other dimensions.
 Two convolutional layers are then used with max pooling, with the expectation that they will find edges,
 and combine them into features, as well as reduce the size of the image down to
 $25\times25$. Figure~\ref{fig:conv} shows the detail of the convolutions.

\subsection{Textline Finding and Reading}
\begin{figure}
\centering
\vspace{-1cm}
\includegraphics[width=\textwidth]{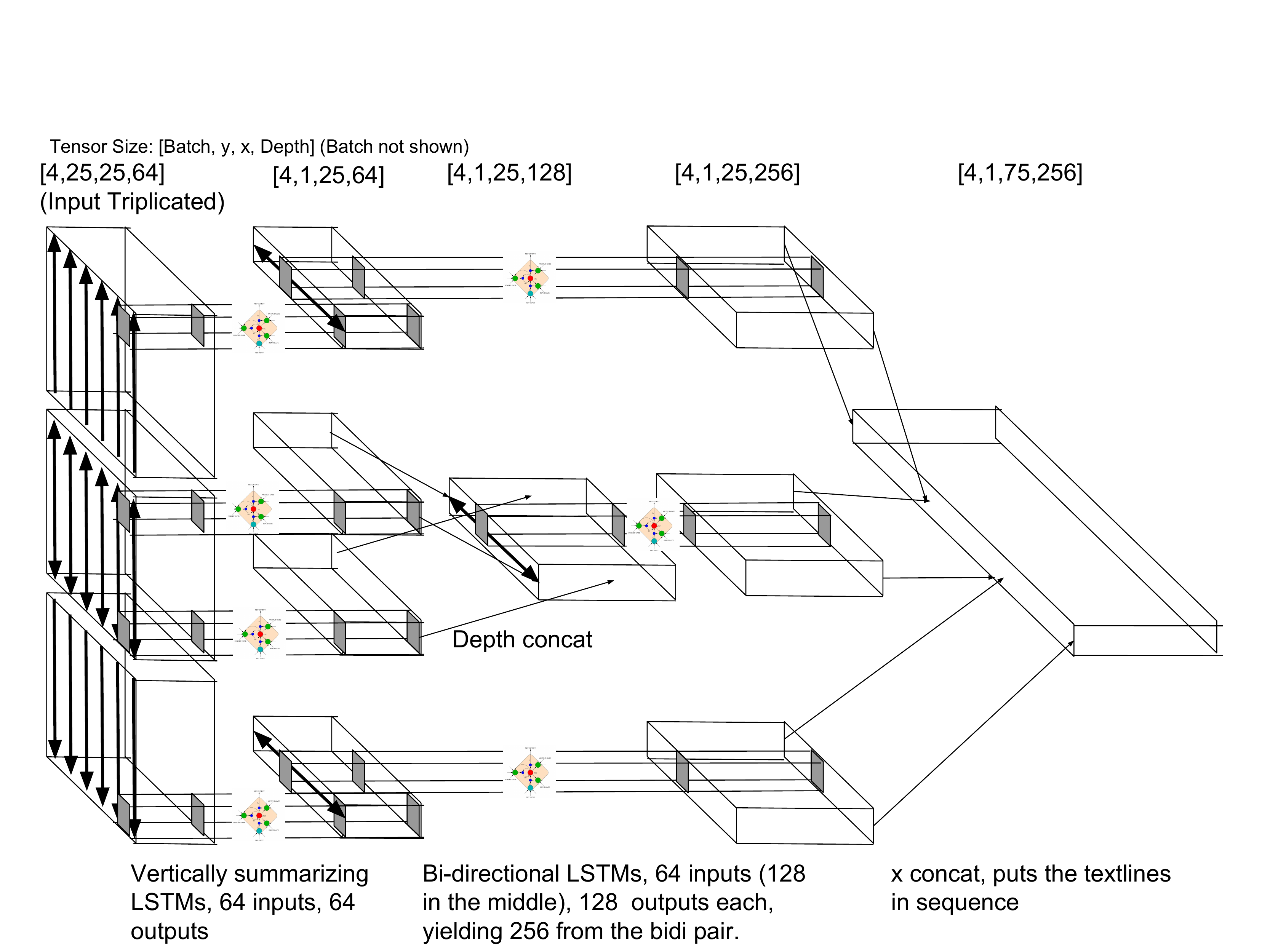}
\caption{Text Line Finding and Reading}
\label{fig:line_find_and_read}
\end{figure}

Vertically summarizing Long Short-Term Memory (LSTM)\cite{hochreiter1997long} cells
 are used to find text lines. \textit{Summarizing} with an LSTM, inspired by the LSTM used for sequence to sequence
translation \cite{NIPS2014_5346}, involves \textit{ignoring the outputs of all timesteps except the last.}
A \textit{vertically} summarizing LSTM is a summarizing LSTM that \textit{scans
 the input vertically.} It is thus expected to
 compute a vertical summary of its input, which will be taken from the last vertical timestep.
 \textit{Each x-position is treated independently.}
 Three different vertical summarizations are used:
\begin{enumerate}
\item Upward to find the top textline.
\item Separate upward and downward LSTMs, with depth-concatenated outputs, to find the middle textline.
\item Downward to find the bottom textline.
\end{enumerate}
Although each vertically summarizing LSTM sees the same input, and could theoretically
 summarize the entirety of what it sees, they are organized this way so that they only have to produce
 a summary of the most recently seen information. Since the middle line is harder to find, that gets
 two LSTMs working in opposite directions. Each receives a copy of the output from the convolutional
 layers and passes its output to a separate bi-directional horizontal LSTM to recognize the text.
 Bidirectional LSTMs have been shown to be able to read text with high
 accuracy \cite{breuel2013high}.
 The outputs of the bi-directional LSTMs are concatenated in the x-dimension, to string the text
 lines out in reading order. Figure~\ref{fig:line_find_and_read} shows the details.

\subsection{Character Position Normalization}

\begin{figure}
\centering
\vspace{-6.5cm}
\includegraphics[width=\textwidth]{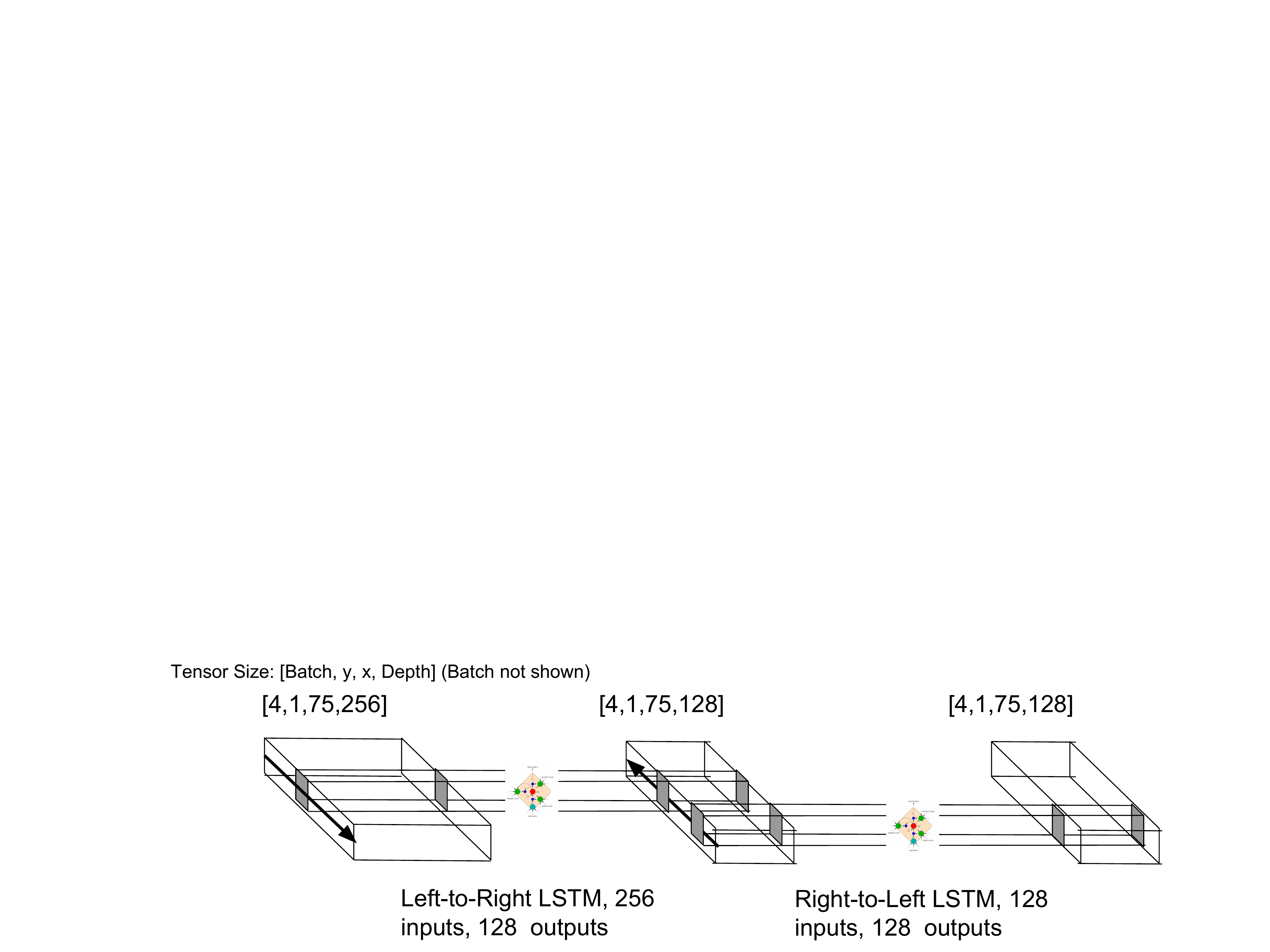}
\caption{Character Position Normalization}
\label{fig:pos_norm}
\end{figure}

Assuming that each network component so far has achieved what it was designed to do, we now have
a batch of four sets of
 one to three lines of text, spread spatially across the x-dimension. Each of the four sign images
 in a batch may have the text positioned differently, due to different perspective within each sign image.
 It is therefore useful to give the network some ability to reshuffle the data along the x-dimension.
 To that end we provide two more LSTM layers, one scanning left-to-right across the x-dimension,
 and the other right-to-left, as shown in Figure~\ref{fig:pos_norm}.
 Instead of a bidirectional configuration, they operate in two distinct layers.
This allows state information to be passed to the right or left in the x-dimension, allowing the characters
in each of the four views to be aligned.

\subsection{Combination of Individual View Outputs}
\begin{figure}
\centering
\vspace{-4.5cm}
\includegraphics[width=\textwidth]{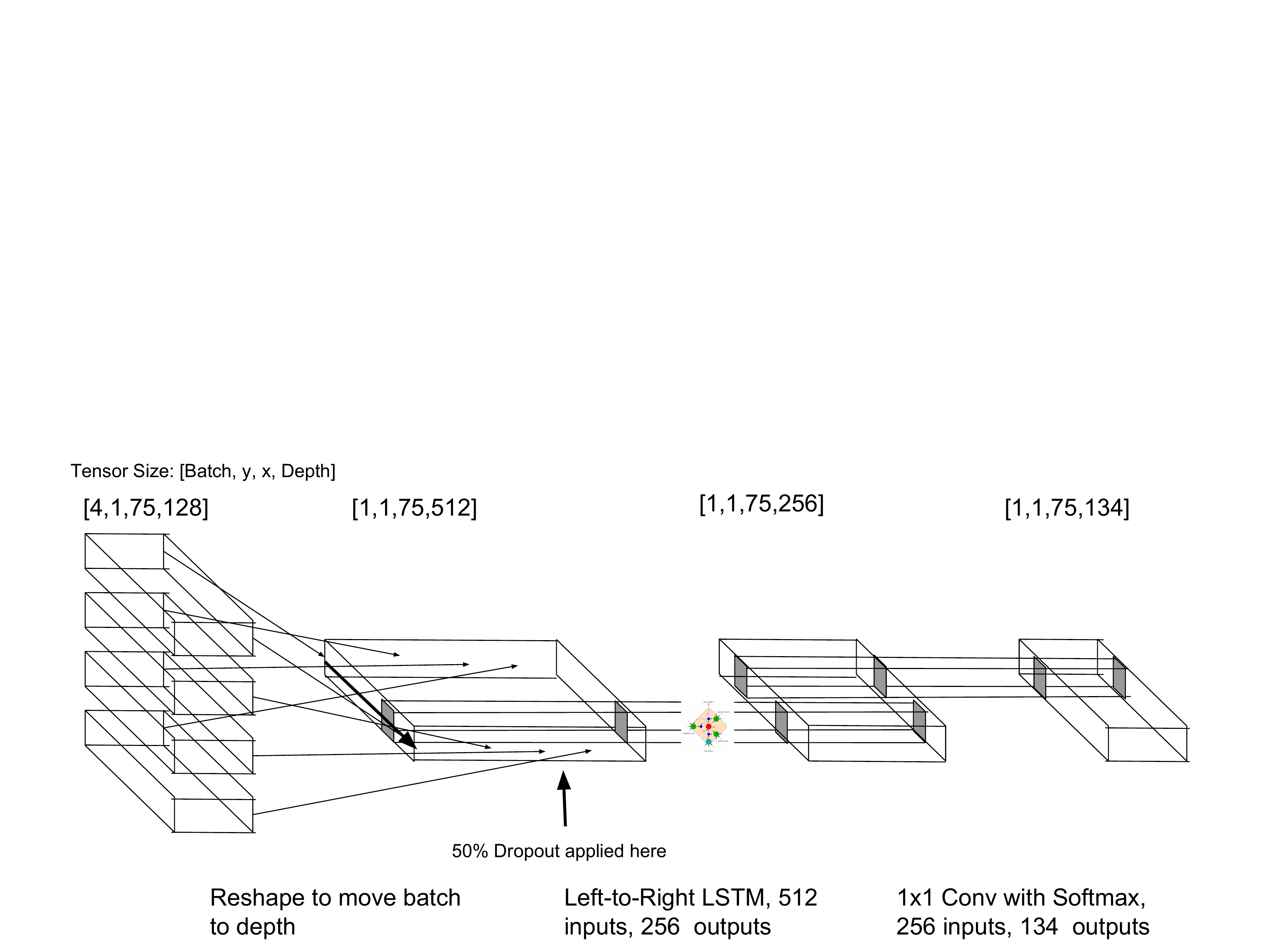}
\caption{Combination of Individual View Outputs}
\label{fig:combinatorial}
\end{figure}

After giving the STREET network chance to normalize the position of the characters along the x-dimension,
 a generic reshape is used to move the batch of 4 views into the depth dimension, which then becomes
 the input to a single unidirectional LSTM and the final softmax layer,
 in Figure~\ref{fig:combinatorial}.
 The main purpose of this last LSTM is to combine the four views fo each sign to produce the most accurate
result. If none of the layers that went before have done anything towards the Title Case normalization,
 this final LSTM layer is perfectly capable of learning to do that well.

The only regularization used is a 50\% dropout layer between the reshape that combines the four signs
 and the last LSTM layer. Details of each component of the STREET graph can be found in
 Table~\ref{table:net_arch}.

\begin{table}
\begin{center}
\caption{Size and computational complexity of the layers in the graph}
\label{table:net_arch}
{\scriptsize
\begin{tabulary}{\linewidth}{| c | C | c | C | c |}
\hline
Name            & Input         & Output       & Weights      & Mult-add \\ \hline \hline 
Reshape0        & 1x150x600x3   & 4x150x150x3  &              & \\ \hline 
Conv0 (5x5x16)  & 4x150x150x3   & 4x150x150x16 & 1216         & 109M \\ \hline 
Maxpool0 2x2    & 4x150x150x16  & 4x75x75x16   &              & \\ \hline 
Conv1 (5x5x64)  & 4x75x75x16    & 4x75x75x64   & 25664        & 577M \\ \hline 
Maxpool1 3x3    & 4x75x75x64    & 4x25x25x64   &              & \\ \hline 
V-SumLSTMs (4x) & 4x25x25x64    & 4x1x25x128x4 & 33024x4      & 330M \\ \hline 
DepthConcat     & 4x1x25x128x2  & 4x1x25x256   &              & \\ \hline 
BidiLSTMs (3x)  & 4x1x25x128x2 + 4x1x25x256 & 4x1x25x256x3 & 263168x2 + 394240 & 92M \\ \hline 
XConcat         & 4x1x25x256x3  & 4x1x75x256   &              & \\ \hline 
LTRLSTM         & 4x1x75x256    & 4x1x75x128   & 197120       & 59M \\ \hline   
RTLLSTM         & 4x1x75x128    & 4x1x75x128   & 131584       & 39M \\ \hline  
Reshape1        & 4x1x75x128    & 1x1x75x512   &              & \\ \hline 
LTRLSTM         & 1x1x75x512    & 1x1x75x256   & 787456       & 59M \\ \hline 
Softmax         & 1x1x75x256    & 1x1x75x134   & 34438        & 2.6M \\ \hline   
Total           &               &              & 2.2M         & 1.3B \\ \hline 
\end{tabulary}
}
\end{center} 
\end{table}

\section{Experiments and Results}
As a baseline, Tesseract \cite{tutorial} was tested, but the FSNS dataset is extremely difficult for it.
 The best results were obtained from the LSTM-based engine in version 4.00, with the addition of
pre-processing to locate the rectangle of the sign, and invert the projective transformation, plus
post-processing to Title Case the output to match the truth text, as well as combination of the highest confidence
results from the four views.
Even with this help, Tesseract only achieves word recall of 20-25\%. See Table~\ref{table:results}.
The majority of failure cases revolve around the textline finder, which includes noise connected components,
 drops characters, or merges textlines. The main cause of these difficulties appears to be the tight line spacing,
compressed characters, and tight border that appears on most signs.

The STREET model was trained using the CTC \cite{graves2006connectionist} loss function,
 with the Adam optimizer \cite{kingma2014adam} in Tensor Flow,
 with a learning rate of $2\times10^{-5}$, and 40 parallel training workers.
The error metrics outlined in Section \ref{challenge} were used.
The results are also shown in
 Table~\ref{table:results}.
 The results show that the model is somewhat over-trained, yet the results for validation, test and private test are very
 close, which suggests that these subsets are large enough to be a good reflection of the model's true
 performance.

\begin{figure}
\centering
\includegraphics[height=2.5cm]{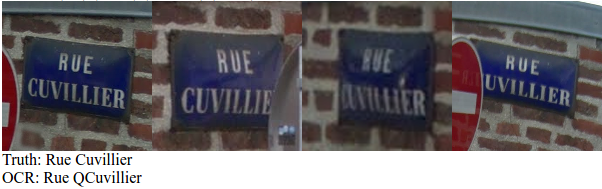}
\includegraphics[height=2.5cm]{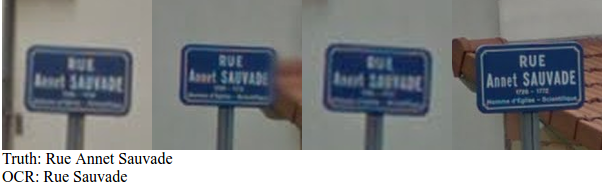}
\includegraphics[height=2.5cm]{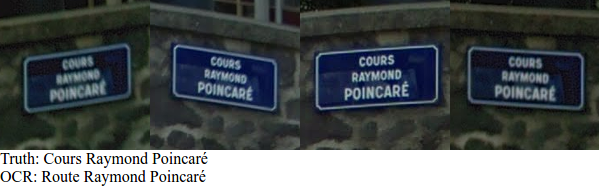}
\includegraphics[height=2.5cm]{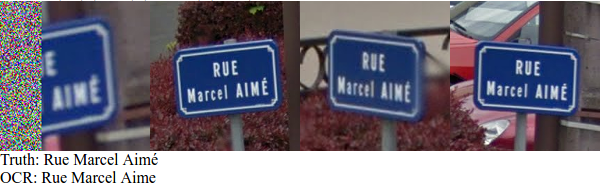}
\caption{Some examples of error cases}
\label{fig:error_cases}
\end{figure}

\begin{table}
\begin{center}
\caption{Error rate results}
\label{table:results}
{\scriptsize
\begin{tabulary}{\linewidth}{| C | C | C | C | C |}
\hline
System & Test Set   & Word Recall & Word Precision & Sequence Error \\ \hline \hline 
Tesseract        & Validation & 22.73       & 20.21          & 95.81 \\ \hline 
Tesseract        & Test       & 23.58       & 20.49          & 98.91 \\ \hline 
Tesseract    & Private Test   & 23.93       & 21.05          & 95.93 \\ \hline 
\hline
STREET        & Train      & 94.90       & 95.40          & 13.14 \\ \hline 
STREET        & Validation & 89.46       & 90.28          & 26.63 \\ \hline 
STREET        & Test       & 88.81       & 89.71          & 27.54 \\ \hline 
STREET    & Private Test   & 89.48       & 90.32          & 26.64 \\ \hline 
\end{tabulary} 
}
\end{center} 
\end{table}
Some examples of error cases are shown in Figure~\ref{fig:error_cases}.
In the first example, the model can be confused by obstructions.
On the second line, the model drops a small word, perhaps as not relevant.
On the third line, a less frequent prefix is replaced by a more frequent one.
In the final example, an accent is dropped.

\section{Conclusion}
The FSNS dataset provides an interesting machine learning challenge. We have shown that it is possible to
 obtain reasonable results for the entire task with a single end-to-end network, and the STREET network
 could easily be improved by application of common regularization approaches and/or changing the network
 structure. Alternatively there are many other possible approaches that involve applying algorithmic or
 learned solutions to parts of the problem. Here are a few:
\begin{itemize}
\setlength\itemsep{0.5em}
\item[\labelitemii] Detecting the position/orientation of the sign by image processing or even structure from motion methods, correcting the perspective, and applying a simple OCR engine.
\item[\labelitemii] Text line finding followed by OCR on individual text lines.
\item[\labelitemii] Detecting the worst sign(s) and discarding them, by blur detection, obstruction
 detection, contrast, or even determining that there is more than one physical sign in the image.
\end{itemize}
A comparison of these approaches against the end-to-end approach would be very interesting and provide useful information for the direction of future research.

\bibliographystyle{splncs03}
\bibliography{fsns}
\end{document}